\documentclass[onecolumn]{article}
\usepackage{xcolor}

\usepackage{amsmath}
\usepackage{authblk}
\usepackage{lettrine}
\usepackage{algpseudocode} 
\usepackage{amssymb}  
\usepackage{amsfonts}
\usepackage{siunitx}
\usepackage{graphicx,subfigure}
\usepackage{enumitem}

\newtheorem{theorem}{Theorem}

\newtheorem{proposition}[theorem]{Proposition}

\begin{document}
\title{On-line learning of dynamic systems: sparse regression meets Kalman filtering}

\author[1]{G. Pillonetto}
\author[2]{A. Yazdani}
\author[3]{A. Aravkin}
\affil[1]{Department of Information Engineering, University of Padova, Padova (Italy)}
\affil[2]{Department of Neurology, University of Texas, Houston (USA)}
\affil[3]{Department of Applied Mathematics, University of Washington, Seattle (USA)}
\date{}                     
\setcounter{Maxaffil}{0}
\renewcommand\Affilfont{\itshape\small}

\maketitle

\begin{abstract}
Learning governing equations from data is central to understanding the behavior of physical systems across diverse scientific disciplines, including physics, biology, and engineering. The Sindy algorithm has proven effective in leveraging sparsity to identify concise models of nonlinear dynamical systems. In this paper, we extend sparsity-driven approaches to real-time learning by integrating a cornerstone algorithm from control theory---the Kalman filter (KF). The resulting Sindy Kalman Filter (SKF) unifies both frameworks by treating unknown system parameters as state variables, enabling real-time inference of complex, time-varying nonlinear models unattainable by either method alone. Furthermore, SKF enhances KF parameter identification strategies, particularly via look-ahead error, significantly simplifying the estimation of sparsity levels, variance parameters, and switching instants. We validate SKF on a chaotic Lorenz system with drifting or switching parameters and demonstrate its effectiveness in the real-time identification of a sparse nonlinear aircraft model built from real flight data.
\end{abstract}


\newpage

\section*{Introduction}

The ability to predict the behavior of physical systems is an essential aspect of modern scientific discovery.  
For dynamical systems, whose behavior is parameterized by a set of differential equations, discovery and prediction fall within the scope of system identification \cite{Astrom71,Soderstrom,Ljung:99}.  
Recent cross-fertilization between system identification and machine learning \cite{Scholkopf01b,SpringerRegBook2022,PillonettoPNAS} leverages regularization theory to obtain parsimonious models that fit available data from a large set of candidate functions in high-dimensional spaces.  
An effective technique to enforce parsimony is to directly penalize the complexity of possible solutions \cite{Tibshirani:96,Bai2019,Bai2018,Rosasco2013,Smith2014,Stoddard2017}.  
In \emph{physics-informed machine learning}, regularizers for dynamical systems incorporate prior knowledge of the underlying physics \cite{Champion2020,Karniadakis2021,nghiem2023,Raissi2019}.  
Along with other algorithms that use, for example, the \(\ell_1\) norm as penalty \cite{tibshirani_regression_1996,PC08,Daub2004}, the Sindy algorithm \cite{Brunton2016,Champion2019} is a leading approach that promotes sparsity to estimate nonlinear systems whose dynamics are described by only a few significant terms.

We provide a powerful generalization of Sindy based on the Kalman filter \cite{kalman1960} (KF), a cornerstone of control theory.  
Proposed in the 1960s, the KF is a statistical processing technique for real-time estimation and prediction of dynamics described by state-space models. One of its earliest applications was the on-board computer that guided the moon landing in 1969 by estimating and predicting the spacecraft's position.  
Since then, the KF and state-space models have become ubiquitous in tracking, navigation, estimation, control, autoregressive modeling, and large-scale data assimilation \cite{Chui2017}.  
This paper develops a model that combines the state-space perspective and KF with Sindy.  
We interpret unknown system parameters as states of a dynamic system, developing a filter for learning parsimonious nonlinear state-space models.

The resulting algorithm, called the Sindy Kalman Filter (SKF), has unique capabilities.  
First, we show that SKF is a true generalization of Sindy, in that under constant dynamics it recovers the same estimates as Sindy. The filtering mechanism extends this foundation to dynamic, noisy, and time-varying settings.  
In particular, SKF provides on-line capabilities for learning nonlinear systems, extending the batch-processing nature of classical Sindy and enabling efficient handling of data streams with thousands of measurements arriving in real time.  
It further incorporates robust and practical approaches to real-time tuning of sparsity using look-ahead prediction methods from filtering theory.  

These features are important in modern engineering, where frequent model adaptation is necessary due to uncertain and evolving dynamics.  
For example, future applications in aviation, including space missions, may require real-time identification to enable autonomous operations such as active monitoring, fault detection, control adaptation, and reconfiguration \cite{DeVivo2014}.  
As a proof of concept, we apply SKF for on-line identification of governing equations in a realistic aircraft model.  
Airplane sensors stream data to an onboard computer, which uses SKF to recursively update the motion equations with real-time control of model complexity, inferring a sparse model with maximal generalization capability.

The dynamic nature of SKF also allows it to estimate time-varying systems \cite{Toth2010}.  
It can learn parsimonious governing equations with parameters that slowly vary over time or be applied to switching systems, a challenging problem in system identification \cite{Sontag1996,JuloskiLN2005,PillonettoHybrid}.  
Such systems describe complex natural phenomena undergoing sudden shifts in dynamics due to internal feedback or changes in external conditions.  
These phenomena include biological systems under non-equilibrium dynamics \cite{Yan2023,Rhind2012,Cadart2018} and bifurcations of normal forms describing spontaneous oscillations in chemical reactions, electrical circuits, and fluid instabilities \cite{Holmes1983}.  
We illustrate this using the Lorenz model, where parameters slowly vary over time or undergo abrupt changes leading to different dynamical regimes \cite{Hassard1981}.

\section*{Sindy Kalman filter}

We begin by reviewing the Sindy algorithm for system identification, 
then describe its on-line extension leading to the Sindy Kalman Filter (SKF).

\subsection*{Sindy}

Consider dynamical systems in state-space form
\begin{equation}\label{StateMod}
\dot{x}(t) = f(x(t))
\end{equation}
where $x(t) \in \mathbb{R}^n$ is the state at instant $t$ and $\dot{x}(t)$ contains the derivatives.
In addition, $f: \mathbb{R}^n \rightarrow \mathbb{R}^n$ is the unknown transition  function with coordinate functions  denoted by $f_i$.\\
The state $x$ is assumed measurable at instants $t_1,t_2,\ldots,t_m$, 
where estimates of derivatives $\dot{x}$ are also available.
To estimate the system, a library of nonlinear functions is used, encoded in a regression matrix $\Theta$
where each column represents a candidate function.
Consider for example the set of monomials up to order $r$ built with the state components, a model closely related to the (truncated) Volterra series \cite{Boyd1985}.  
If $r=2$ and the state dimension is $n=2$, the $i$th row of $\Theta$ is given by 
\begin{equation}
\Theta_i=\left[\begin{array}{ccccc} x_1(t_i) & x_2(t_i)  &  x^2_1(t_i) & x^2_2(t_i) & x_1(t_i)x_2(t_i) \end{array}\right].
\end{equation}
To simplify notation, we focus w.l.o.g. on estimation of a single component $f_i$ of the transition function
and write the measurements model as
\begin{equation}\label{MeasModY}
y = \Theta \Xi + e
\end{equation}
where the column vector $\Xi$ contains the monomial expansion coefficients while
the vector $y$ contains the derivatives of $x_i(t)$ corrupted by the noise $e$. 
The dimension of $\Xi$ grows rapidly 
with $r$ and $n$ but 
in many natural physical systems only a few of the columns, in this case monomials, are truly active.
Sparsity of the vector $\Xi$ is used to both control model complexity and to learn interpretable models.
Sindy estimates the motion equations from $y$ by solving a sparsity-promoting least squares problem,
where the sparsity parameter $\lambda$ controls how many system components are set to zero. 
Such parameter can be estimated through cross-validation, minimizing the prediction error on validation data not contained in $y$.
The Sindy algorithm is summarized in the pseudocode reported in Appendix.\\

\subsection*{Sindy Kalman filter}

We now consider more complex dynamical systems in state-space form
where the governing equations may vary in time, i.e.
\begin{equation}\label{StateMod2}
\dot{x}(t) = f(x(t),t).
\end{equation}
The generalization allows the parameter vector
$\Xi$ to depend on $t$, and we can consider the observations to be arriving in real time. The model for the measurements  $y_t$ is given by 
\begin{equation}
y_t = \Theta_t  \Xi_t + e_t, \quad t=1,2,\ldots.
\end{equation}
However, it is clear that if we allow all the components of $\Xi$ to be independent of each other, and
use e.g. a least squares criterion to determine them, the data can be perfectly interpolated. The resulting model
is not interpretable and has no prediction capability on future outputs. For this reason,
various tracking and adaptive algorithms have been proposed in the literature \cite{And1985,Gust2001,Ohl2010}.
We consider the problem from a different perspective.
To deal with intrinsic ill-posedness, we set up a model to describe 
relationships between the parameter vectors, and include 
sparsity information on the $\Xi_t$
in the estimation process. \\ 
For many natural systems, 
 one expects similar values
for $\Xi_t$ that are temporally close each other, except at switching instants (e.g. for regime changes). 
In a Bayesian framework, 
we therefore aim to control complexity 
by modeling the $\Xi_t$ as a sequence of correlated random vectors, which vary smoothly in time when the correlation is large. 
In the Kalman framework, this is done by 
taking $\Xi_t$ as the states and $y_t$ as observed outputs of the following stochastic state-space model:
\begin{eqnarray} \label{Kmodel}
\Xi_{t+1} &=& A \Xi_{t} + w_t \\ \nonumber
y_t &=& \Theta_t  \Xi_t + e_t.
\end{eqnarray}
The dynamics of $\Xi$ are regulated by the transition matrix $A$ and driven 
by uncorrelated zero-mean noises $w_t$ with covariance matrix $Q_t$. 
The system initial condition $\Xi_1$ is a random vector of mean $\mu$ 
and covariance matrix $P$, independent of the noise. 
Later on, it will be seen that specific choices of the matrices $A,Q_t$ and $P$
provide connections with Sindy 
and the different classes of time-varying nonlinear systems mentioned in the introduction.\\
As new data become available, recursive sparse estimates of the $\Xi_t$ are generated by 
complementing the classical Kalman equations 
with a Sindy-type procedure with sparsity parameter $\lambda$.
This defines the SKF illustrated in Fig. \ref{FigSKF} whose
pseudocode and properties are illustrated in Appendix. In that section we also discuss approaches for real-time estimation of unknown parameters
that may be present in the filter without using any validation data. 
Relying only on the measurements $y_t$, SKF can  in fact estimate
the sparsity level $\lambda$, 
entries of the covariances $Q_t$ or switching instants where
$\Xi_t$ has an abrupt change. 
This is made possible by the fact that at each time step, SKF also returns the one-step ahead error, an important prediction concept in system identification \cite{Ljung:99}. It corresponds to the difference between the predicted output value at time $t$, calculated using only the data available up to $t-1$, and the actual observed value when it becomes available. The mean 
of the squared errors defines what is called in the following the average prediction error, and the parameter estimates minimize it.

 \begin{figure*}
 {\includegraphics[scale=0.8]{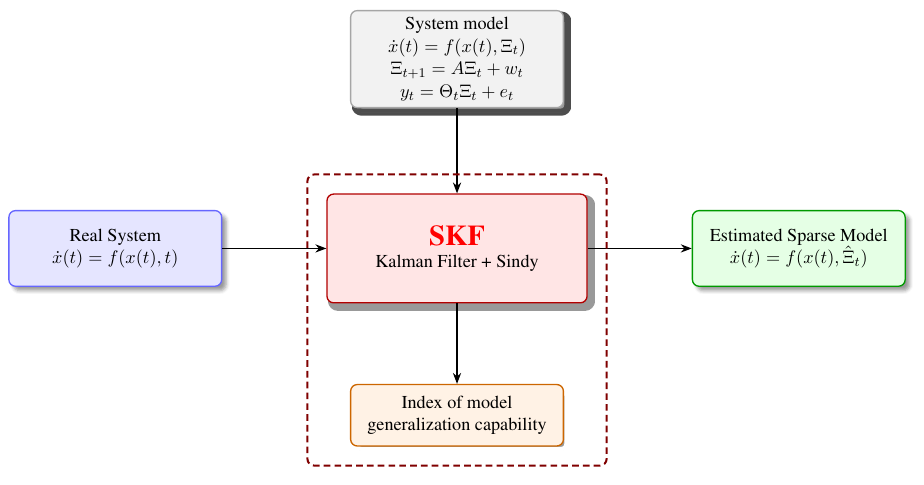}} 
 \caption{\textbf{The left block} represents a nonlinear, time-varying dynamical system with state vector $x(t)$, evolving according to an unknown function $f$. An experiment is performed on this physical system to collect training data $\{x_t, y_t\}$, from which $f$ is to be estimated.
\textbf{The top block} depicts the \textit{system model}, which approximates the behavior of the real system. It includes unknown parameters $\Xi_t$, which may vary over time and typically represent coefficients of monomials, such as in a truncated Volterra series. 
They are viewed as latent states in a higher-level state-space model and are linearly related to noisy measurements $y_t$ through regression matrices $\Theta_t$ built from $x_t$.
\textbf{The central block} shows the \textbf{SINDY Kalman Filter (SKF)}, which estimates $\Xi_t$ in real time by combining Kalman filtering and sparse identification (SINDY). This promotes sparsity and extracts the most relevant components of $\Xi_t$ that govern the system's dynamics.
\textbf{The right block} receives the estimated parameters $\hat{\Xi}_t$ to reconstruct a sparse model capable of replicating the original system's behavior. \textbf{The bottom block} highlights a real-time model generalization index (e.g., one-step-ahead prediction error), which can help infer hidden variables or switching behavior, and enhance model structure selection.
As a result, SKF provides sparse, interpretable models with strong predictive capability for future (unseen) system behavior. Thanks to its linear computational complexity with respect to the data set size, the SKF structure is efficient and also applicable to large-scale batch identification problems.
}
 \label{FigSKF}
\end{figure*} 

\section*{Results}

\subsection*{Three on-line learning scenarios illustrated with the Lorenz system}

To show the potentiality of SKF,
we consider three situations
related to time-invariant, switching and smoothly changing
dynamic systems.  
In each situation 
we use as case study the chaotic Lorenz system,
one of the first Sindy applications 
documented in \cite{Brunton2016}.
The model is
\begin{eqnarray}\label{Lorenz}
\dot{x}_1&=&\sigma(x_2-x_1),\\ \nonumber
\dot{x}_2&=&x_1(\rho-x_3)-x_2,\\ \nonumber
\dot{x}_3&=&x_1x_2-\beta x_3.
\end{eqnarray}
and give rise to rich and chaotic dynamics evolving on an attractor.
One can see that the system is sparse, there are only a few terms in the right-hand side of each equation. 
We illustrate numerical results from single noise realizations. They are well representative
of the average performance of the estimator which has been assessed also using Monte Carlo studies.

\subsubsection*{Time-invariant system and connection with Sindy}

In the time-invariant setting, SKF is defined by the correspondences $A=I$ and $Q=0$, so that the parameter dynamics reduce to $\Xi_{t+1} = \Xi_{t} =: \Xi$.  
The covariance matrix $P$ describes our information on $\Xi$ prior to observing any data.  
If we let $P = \gamma I$ and make $\gamma$ grow to infinity, we assign a non-informative prior to the filter, indicating no prior knowledge of the dynamic system parameters before observing the states.  
As discussed in Appendix, such a prior makes the filter estimate independent of the noise variance, and the following result holds.\\

\emph{In the time-invariant case, if the prior on the system parameter vector $\Xi$ is non-informative, Sindy and SKF return the same sparse estimates. If the prior is informative, under mild assumptions the same result holds asymptotically.}\\

Hence, in the setting originally treated in \cite{Brunton2016}, SKF is perfectly equivalent to Sindy, with the added advantage that it can be used in an on-line context as well.  
This means governing equations can be updated incrementally as new data arrive, rather than recomputed from scratch. \\
 
\begin{figure*}
 {\includegraphics[scale=0.75]{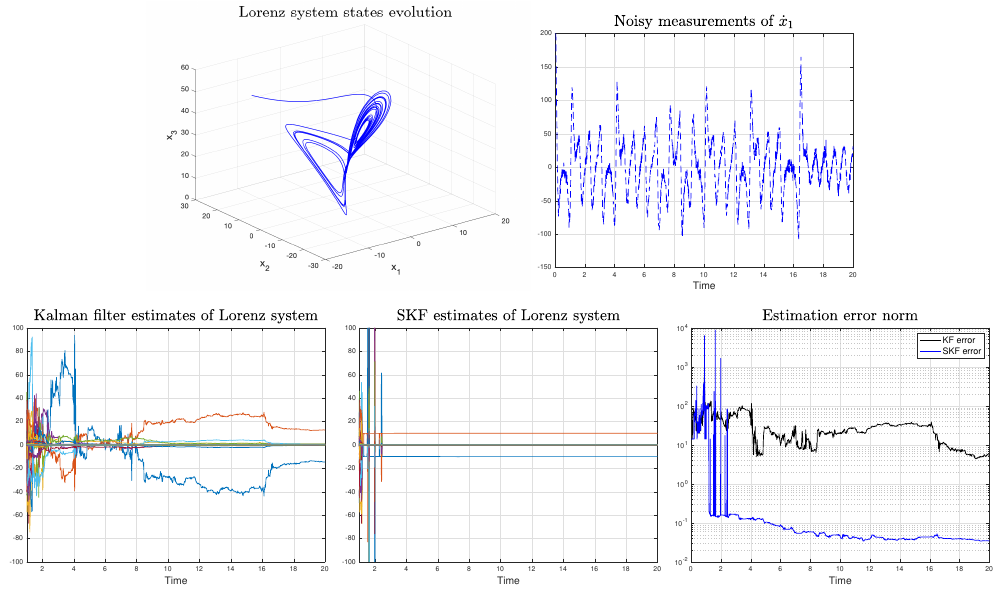}}  
\caption{Identification of a time-invariant Lorenz system.
The example consider estimation of the first governing equation $\dot{x}_1=\sigma(x_2-x_1)$,
with $\sigma=10$. So, there are just two parameters different from 
zero, equal to $10$ and $-10$. 
The evolution of the system is shown in the top left panel while 2000 noisy measurements of the first 
governing equation are in the top right. The signal-to-noise ratio (SNR), expressed in decibels, is around 15 dB.
The unknown parameters of the Lorenz system are 
the coefficients of monomials up to $4$th order and are  
the state of a Kalman filter. They are estimated in a on-line manner as noisy derivatives become available in time.
Classical Kalman estimates are in the bottom left panel. No sparsity is promoted, so that all the estimates
are different from zero at any instant. 
Their behaviour is somewhat erratic. At the end of the experiment 
the estimated value of the nonzero monomials coefficients is $-14.4,12.7$.
SKF estimates are in the middle panel. The new procedure
tries to determine at any iteration the most predictive sparse model
by minimizing the average one-step ahead error.
After few steps the governing equation returned by the filter is very close to truth
and a perfect sparsity pattern is reconstructed. At the end of the experiment,
the estimated value of the nonzero monomials coefficients are $-10.01,9.97$.
The time-course of the norm of the estimation error of the two filters is reported in the bottom right panel. }
\label{ExA}
\end{figure*}

We demonstrate SKF’s performance using the Lorenz system with parameters $\sigma=10$, $\beta=8/3$, and $\rho=28$, focusing on the first governing equation $\sigma(x_2 - x_1)$.  The candidate functions are monomials up to the 4th order.  During each unit time, 100 data points become available.  
Results reported in Fig.~\ref{ExA} show the time evolution of the system parameter estimates, illustrating how SKF dramatically improves classical Kalman filter estimates by exploiting sparsity.  The sparsity pattern of the first Lorenz equation is perfectly reconstructed on-line shortly after time instant 3, when the filter has processed just over 300 measurements.  Once the correct pattern is recovered, the filter's ability to provide minimum variance estimates is activated, as discussed in the Methods section.  
SKF then produces statistically optimal estimates, i.e. no alternative method has a lower mean squared error. Consequently, by the end of the experiment, the estimated parameters are very close to the true values, with a relative error of approximately $0.3\%$.

\subsubsection*{Smoothly changing systems}

In this second setting, some system parameters are expected to vary smoothly over time while many others may remain constant or be effectively null.  
In many applications, temporal variation reflects gradual physical or structural changes in the underlying system. For example, in adaptive control and on-line system identification, parameter dynamics are often modeled as stochastic processes to capture effects such as changing payloads or friction \cite{Ljung:99}. In structural health monitoring and civil engineering, parameters such as stiffness or damping ratios are treated as random to model the impact of aging, fatigue, or environmental conditions \cite{Sohn2003, Zonta2004, Fassois2007}.  
Our framework allows describing such evolution via a linear stochastic process, $\Xi_{t+1} = A \Xi_{t} + w_t$, where the choice of the transition matrix $A$ and the noise covariances $Q_t$ can be driven by domain knowledge about the expected smoothness, reversibility, and timescale of parameter evolution.\\

When the matrix \( A \) is stable, i.e., all its eigenvalues lie within the unit circle, the parameter trajectories remain bounded in probability and tend to exhibit mean-reverting behavior. A simple example is \( A = a I \) with \( |a| < 1 \), which produces a first-order autoregressive AR(1) process for each component of \(\Xi_t\). Each parameter is then pulled toward zero with dynamics regulated by \( a \), limiting long-term drift. A nonnegative \( a \) induces positive correlation, yielding smooth trajectories. Importantly, the state vector $\Xi_{t+1}$ need not represent only the physical parameters of interest. It can be augmented with latent or auxiliary variables that do not directly appear in the regression model but help capture more complex temporal dynamics. This expanded state-space formulation allows modeling any stationary process with a rational spectral density, as ensured by Lyapunov theory and classical system identification principles \cite{Ljung:99,DurbinKoopman2012}.\\

If \( A = I \), the process is a random walk, allowing parameters to drift freely over time. This model connects closely to splines \cite{Wahba:90,deBoor2001} and offers a simple way to represent regular functions without strong prior assumptions. It is the most used in applications, e.g., in econometrics and finance to capture evolving relationships like inflation or volatility \cite{DurbinKoopman2012}, or in neuroscience to describe gradual adaptation in neural encoding \cite{Pillow2008}.  To illustrate this, consider the Lorenz system where the parameter \( \sigma \) follows the continuous trajectory (dashed line) shown in the right panel of Fig. \ref{ExC}. After time 3, it decreases linearly from 20 to 10. As in the previous examples, we use monomials up to order 4 to describe system dynamics. Our information is that the linear part of the system could smoothly change in time. Hence, we describe the related monomial coefficients as independent random walks of common variance, estimated by minimizing the average prediction error. The other coefficients are unknown but assumed constant. Fig. \ref{ExC} compares estimates from the classical Kalman filter and SKF. Also in this setting, the right panel illustrates how the proposed physics-informed filter can dramatically improve the system identification procedure.

{
 \begin{figure*}
{\includegraphics[scale=0.24]{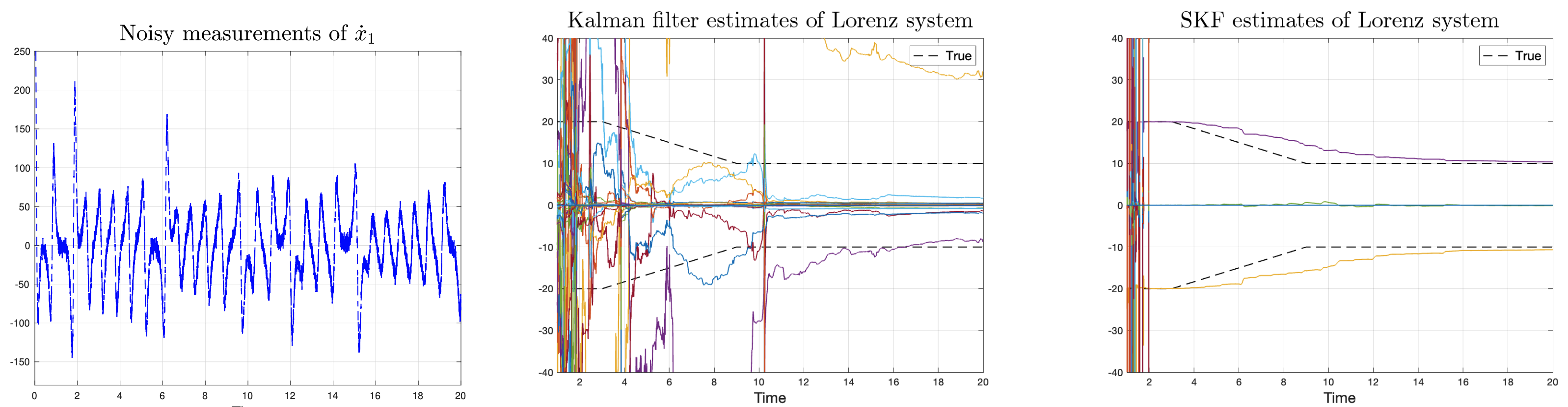}}
\caption{Identification of a smoothly time-varying Lorenz system.
The first governing equation is $\dot{x}_1=\sigma(t) (x_2-x_1)$,
where the time-course of  $\sigma$ is the black dashed line in the right panels. 
After the time instant 3, its decreases linearly from 20 to 10.
The left panel shows 10000 noisy measurements of $\dot{x}_1$,
the SNR is around 15. As in the previous case studies
the unknown system parameters are 
the coefficients of monomials up to $4$th order. 
Parameters of the linear part of the system are modeled as independent random walks 
whose common variance is estimated by minimizing
the average prediction error. 
Estimates from the classical Kalman filter 
are in the middle panel, they 
are unable to track the evolution of Lorenz parameters.
SKF estimates are in the right panel, obtained with the sparsity parameter $\lambda=0.4$.
An estimate with the right level of sparsity is obtained. Furthermore, SKF learns from data 
that the value of $\sigma$ decreases: its estimate is close to 20 immediately after $t=2$ and
then converges to 10 as time progresses.}
\label{ExC}
\end{figure*}
}

 \begin{figure*}
\center {\includegraphics[scale=0.8]{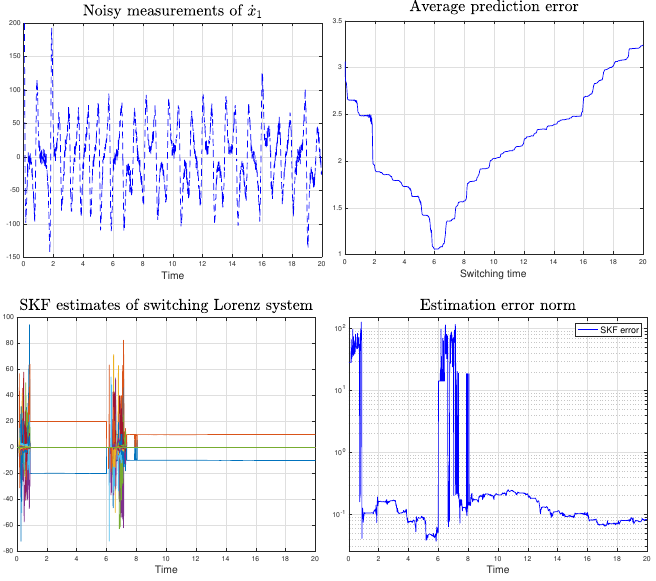}} 
\caption{Identification of a switching Lorenz system.
In this example the parameter $\sigma$ of the first governing equation $\dot{x}_1=\sigma(x_2-x_1)$
follows an abrupt change, moving from the value $20$ to $10$ at time instant $t=6$.
The 2000 noisy measurements (SNR around 15) of the first 
governing equation are in the top left panel. One can see that
it is difficult to detect from them the instant where 
the dynamics shift happens.
The unknown parameters of the Lorenz system are 
the coefficients of monomials up to $4$th order and are  
the state of SKF. Estimate of the switching instant is the minimizer of the average prediction error (see also Appendix for details).
The resulting objective is displayed in the top right panel and its minimizer is exactly equal to $t=6$. 
SKF outcomes are in the bottom left panel and are obtained setting the switching instant to its estimate.
The governing switching equation returned by the filter is very close to truth
and a perfect sparsity pattern is obtained. Just before the switching instant and at the end of the experiment
the estimated values of the nonzero monomials coefficients are $-19.94,20.00$ and $-9.93,9.95$ .
The time-course of the estimation error norm is visible in the bottom right panel.}
\label{ExB}
\end{figure*}

\subsubsection*{Switching systems}

This third setting concerns a special situation in which the system parameters evolve according to a piecewise constant time course. Changes occur only at the instants $\tilde{t}_k$, which are relatively rare.  
The SKF is defined by the correspondences $A = I$ and $Q = 0$ at all time instants, except immediately before $\tilde{t}_k$, where $A = I$ and $Q = \gamma I$ with a large $\gamma$. In practice, at $\tilde{t}_k$ the filter is re-initialized with a non-informative prior.  
The switching instants and the sparsity parameter are unknown and are estimated by optimizing the predictive scores computed by the SKF. 
The estimation is performed \emph{on-line} by running in parallel a bank of filters, each associated with a different hypothesized switching time. At each time instant, the result is provided by the filter whose predictive score attains the highest value. This strategy allows for real-time localization of switching points without requiring a priori knowledge of their position.\\

We consider a Lorenz system undergoing a dynamic shift at time $t = 6$. Specifically, the parameter $\sigma$ follows a piecewise trajectory, moving from its initial value of $20$ to $10$. A bank of $40$ filters are run in parallel. Results are shown in Fig.~\ref{ExB} and show that the model is perfectly segmented.  
SKF is effective because, when the hypothesized switching time $t_k$ deviates from the true switching instant, the predictive score quickly deteriorates: the filter attempts to forecast future data using a (sparse) model with incorrect parameters. This sharp drop in predictive performance --- visible in the top-right panel of Fig.~\ref{ExB} --- enables accurate localization of switching points.  
Once the correct switching instant is detected, the optimality properties of the SKF, previously discussed, can be fully exploited. Immediately before the switching instant and at the end of the experiment, parameter estimates exhibit an error not larger than $0.7\%$.

{
 \begin{figure*}
{\includegraphics[scale=0.75]{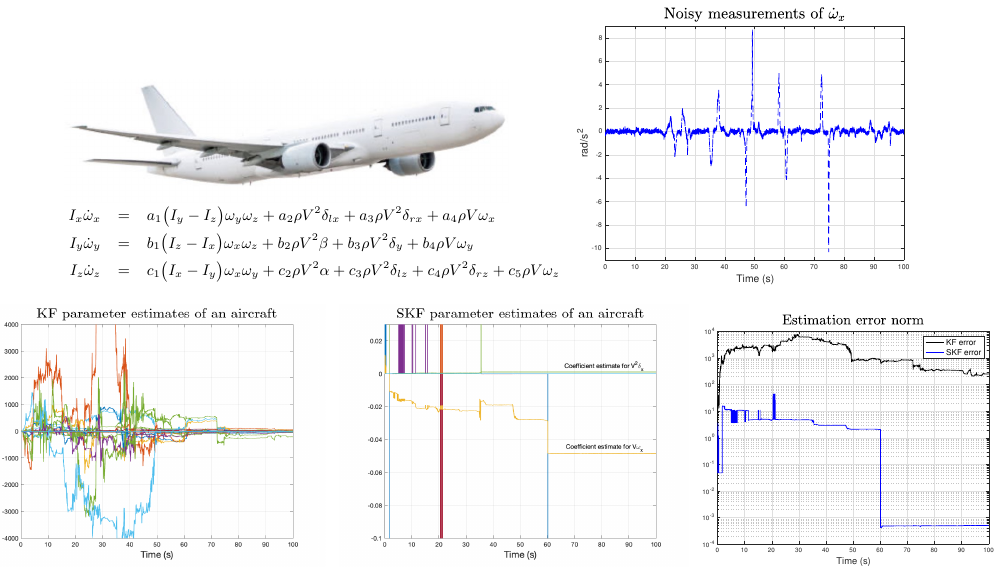}} 
\caption{Identification of an aircraft model. 
The three-dimensional nonlinear motion equations describing the angular velocities of the shaft are in the top left panel.
We consider  estimation of the first equation  $\dot{\omega}_x = \num{1.061e-3} V^2 \big(\delta_{lx}-  \delta_{rx}\big) \num{-4.8e-2}  V\omega_x$, derived from real data in \cite{Li2023}, from the 10000 noisy derivatives  in the top right panel.
In this example, such derivative estimates are obtained by fitting smoothing splines \cite{Wahba:90} to noisy measurements of angular velocities (with an SNR of approximately 20) over a moving temporal window of 1 second, containing 100 measurements. This window is updated every 0.5 seconds.
Basis functions given by monomials up to third-order are built using the following signals measurable by sensors in the aircraft: angular velocities $\omega_x,\omega_y,\omega_z$, aileron angle $\delta_{lx}-\delta_{rx}$ and velocity $V$. Thus, the model depends only on two monomials, one of second- and the other of third-order. In this thought experiment the on-board computer of the aircraft  estimates in real-time the motion equation by making run in parallel a bank of SKF filters equipped with different sparsity  parameters $\lambda$. 
At any time instant, 
the estimate of $\lambda$ is obtained as the minimizer of the average of the prediction errors computed up to that instant. 
The classical Kalman estimates are in the bottom left panel. The filter does not have a good control of complexity since no sparsity can be promoted. Monomial coefficients estimates are different from zero at any instant and far from the true values. 
The middle panel instead shows how SKF improves dramatically model quality as time advances. After 60 seconds  
the right sparsity pattern is found: SKF learns from data that velocity enters the model both linearly and quadratically in combination with other two state variables. At the end of the experiment the estimated equation is 
$\dot{\omega}_x = \num{1.07e-3} V^2 \big(\delta_{lx}-  \delta_{rx}\big) \num{-4.83e-2}  V\omega_x$.}
\label{ExAir}
\end{figure*}
}

\subsection*{On-line identification of an aircraft model}

Accurate aerodynamic models are crucial to achieve aircraft control objectives such as landing precision and relative position accuracy \cite{Maine1985,Jat2015}.  
In real applications, these models may require frequent on-board adaptation due to uncertain and evolving dynamics, possibly caused by damage, which can render governing equations derived from ground-based experiments obsolete.  
Advances in this field promise to enable new guidance and navigation systems for stabilization and tracking.  
For this reason, as a proof of concept, we address a problem where a realistic aircraft model must be estimated to obtain updated in-flight plant knowledge.  
Differently from previous approaches such as \cite{Majeed2018,Amin2019,Wu2021,Verma_Peyada_2021}, we perform sparse on-line system identification for the first time to derive highly interpretable motion equations of the aircraft, relying on very limited aerodynamic information.\\

Pioneering work on modeling and identification of aircraft flight dynamics via least squares can be found in \cite{Greenburg1951,Shinbrot1951}.  
Here, we adopt the classical deterministic model described also in \cite{Haiquan2022}, consisting of the three-dimensional nonlinear motion equations displayed in the top left panel of Fig.~\ref{ExAir}.  
The scalars $I_x, I_y, I_z$ denote the inertia moments, $\rho$ is the air density, and the $a_i$ are unknown parameters.  
The model also includes measurable signals from onboard sensors: angular velocities of the shaft $\omega_x, \omega_y, \omega_z$, aileron/rudder/elevator angles $\delta_x, \delta_y, \delta_z$, angles of attack and sideslip $\alpha, \beta$, and velocity $V$.  
To simulate the angular velocities, we use measurements taken during a real flight and reported in \cite{Li2023} (see also Appendix for details).  
This aspect is important for testing SKF since, in operational settings, some classes of excitation signals are not allowed due to high risks, so only data coming from the system’s natural operational excitation can be used.  
Model parameters from \cite{Li2023} then lead to the following system:

{{\small
\begin{eqnarray} 
\dot{\omega}_x &=& \num{1.061e-3} V^2 \big(\delta_{lx}-  \delta_{rx}\big) \num{-4.8e-2}  V\omega_x\\ \nonumber
\dot{\omega}_y &=& -\num{1.95e-6} \omega_x \omega_z + \num{3.5e-5} V^2 \beta \num{-2.1e-5} V^2 \delta_{y} -\num{2.5e-3}  V \omega_y\\ \nonumber
\dot{\omega}_z &=&  \num{1.6e-6}  \omega_x \omega_y +\num{ 1.8e-5} V^2 \alpha +   \num{1.5e-6} 
V^2 \big(\delta_{lz} +\delta_{rz}\big)  -\num{5.1e-4} V \omega_z.
\end{eqnarray} 
}
We focus on the first equation describing the evolution of the angular velocity \(\omega_x\), which, as also noted in \cite{Li2023}, is the most stressed during flight, primarily due to rolling induced by the pilot's control. The other two signals, \(\omega_y\) and \(\omega_z\), instead exhibit only minor fluctuations.  
On-line sparse identification is performed without assuming any significant prior knowledge about the airplane dynamics. The states and signals used to define monomials up to third order include \(\omega_x, \omega_y, \omega_z\), the differential aileron angle \(\delta_{lx} - \delta_{rx}\), and the velocity \(V\). Consequently, the first equation depends only on a second- and a third-order monomial. These terms must be estimated from the 10,000 noisy measurements shown in the top right panel of Fig.~\ref{ExAir}.\\

Identification is performed in real-time using a single SKF filter. 
As new data arrive, the aircraft parameters and the sparsity parameter \(\lambda\) are updated by minimizing the prediction errors.  
Results in the bottom panels of Fig.~\ref{ExAir} demonstrate that SKF substantially outperforms the classical Kalman filter, which lacks adequate control over model complexity and is unable to promote sparsity. Instead, SKF identifies the correct sparsity pattern immediately after 60 seconds.  
All parameters are then set to zero except for the coefficients related to the two monomials governing the aircraft motion.  
At the end of the experiment, the estimated equation closely matches the true one, given by
\[
\dot{\omega}_x = \num{1.07e-3} V^2 (\delta_{lx} - \delta_{rx}) + \num{-4.83e-2} V \omega_x.
\]

These outcomes highlight some important features of SKF. As already mentioned, the system operates in a closed-loop configuration, where the pilot controls the aircraft by adjusting shaft velocities and angular velocities. This setting presents a significant challenge, as the resulting inputs are typically poorly exciting, leading to data that may not be particularly informative for identifying the full aircraft dynamics. In such cases, there is a risk that the estimated model is accurate only within certain frequency bands, specifically those relevant to the currently applied control. Moreover, closed-loop data often give rise to ill-conditioned estimation problems. The standard Kalman filter struggles under these conditions, which explains the poor quality of its estimates. In contrast, SKF’s ability to produce on-line sparse estimates proves crucial. By isolating only a few key regressors, SKF can accurately identify the entire model even from limited or low-excitation data. As a result, the full dynamics of the aircraft are reconstructed on-line with high precision.

\section*{Conclusions}

We have developed a powerful SKF framework to learn nonlinear dynamical systems from data. It generalizes prior work, including the popular Sindy procedure as a special case. A key advantage of SKF is its ability to perform online identification. In particular, this allows sparse estimation of governing equations to be updated incrementally as new data arrive, rather than being recomputed from scratch, resulting in computational efficiency gains as well as a paradigm shift in modeling.\\ 

Another significant benefit of SKF is its capability to identify time-varying systems, with parameters that change either slowly or abruptly. The latter includes switching systems, which are important in many fields such as gene expression~\cite{Rhind2012,Cadart2018}, population dynamics, and biological systems under non-equilibrium conditions~\cite{Yan2023}. The synergy between Sindy and the Kalman filter, achieved through SKF, also enables us to leverage key innovations from control theory in identifying governing dynamics. A Kalman filter provides real-time measures of model predictivity, notably the one-step ahead prediction error, a well-established criterion in system identification~\cite{Ljung:99}. Optimizing this criterion allows for automatic selection of variances related to stochastic parameter trajectories, switching instants and sparsity parameters in the governing equations. Numerical examples presented here demonstrate that models derived via SKF are both highly interpretable and generalizable.\\

In conclusion, SKF represents an important step forward in physics-informed machine learning for dynamic systems. Regarding future work, the state-space sparse model describing parameter evolution could be enriched by integrating concepts from optimization and control theory, including state-space constraints and robustness to outliers.


\section*{Appendix}

\subsection*{The SKF algorithm}

The Sindy pseudocode is reported below.\\

{\bf{\large Sindy}}
\begin{algorithmic}[1]
\State $\hat{\Xi}= \big(\Theta^\top \Theta\big)^{-1} \Theta^\top y$ \Comment{Least squares estimate}
\State  Until convergence do 
\State  \indent  smallinds = $|\hat{\Xi}|<\lambda$; \Comment{find small coefficients}
\State  \indent  $\hat{\Xi}(\text{smallinds})=0$; \Comment{and threshold}
\State  \indent  biginds = $|\hat{\Xi}| \geq \lambda$; \Comment{find coefficients significant w.r.t. $\lambda$}
\State  \indent  $\widetilde{\Theta}= \Theta(:,\text{biginds})$ \Comment{regression matrix restricted to significant terms}
\State  \indent  $\hat{\Xi}(\text{biginds}) = \big(\widetilde{\Theta}^\top \widetilde{\Theta} \big)^{-1} \widetilde{\Theta}^\top y$;
\Comment{regress dynamics onto remaining terms to find sparse $\Xi$}
\State \Return estimate $\hat{\Xi}$ 
\end{algorithmic}

\medskip

\noindent SKF is instead described by the following pseudocode where 
$\eta^2$ is used to denote the noise variance.\\

{\bf{\large Sindy Kalman filter}}
\begin{algorithmic}[1]
\State $\hat{\Xi}_{1|0}=  \mu$ \Comment{Prediction at $t=1$}
\State $P_{1|0} = P$   \Comment{Predicted error covariance at $t=1$} 
\State {\bf{Recursive update of estimates for $t=1,2,3,\ldots$}}
\State $e_t=y_t -  \Theta_t  \hat{\Xi}_{t|t-1}$   \Comment{Innovation}
\State $V_t = \Theta_t P_{t|t-1}  \Theta_t^\top + \eta^2$ \Comment{Innovation variance}
\State $\hat{\Xi}_{t|t} = \hat{\Xi}_{t|t-1} + P_{t|t-1}  \Theta_t^\top V^{-1}_t e_t$  \Comment{Filtered estimate at $t$}
\State $P_{t|t} = P_{t|t-1} - P_{t|t-1}  \Theta_t^\top V^{-1}_t \Theta_tP_{t|t-1}$  \Comment{Filtered error covariance at $t$} 
\State $\hat{\Xi}_{t+1|t} = A \hat{\Xi}_{t|t}$  \Comment{Prediction at $t+1$}
\State $P_{t+1|t} = AP_{t|t-1}A^\top +Q_t$  \Comment{Predicted covariance error at $t$} 
\State Set $\hat{\Xi}_{t}= \hat{\Xi}_{t|t}$ and until convergence do
\State \indent smallinds = $|\hat{\Xi}_{t}|<\lambda$; \Comment{find small coefficients}
\State \indent $\hat{\Xi}_{t}=\hat{\Xi}_{t|t} - P_{t|t}(:,\text{smallinds}) \big(P_{t|t}(\text{smallinds},\text{smallinds})\big)^{-1} \hat{\Xi}_{t|t}(\text{smallinds})$; \Comment{Sparse estimate at $t$}
\State \Return estimate $\hat{\Xi}_t$ 
\end{algorithmic}

\medskip 

We now briefly discuss some properties of SKF and strategies for estimating possible unknown parameters present in the filter. The reader is referred to the second part of this Appendix for full details.\\
The algorithm reported above shows that SKF exploits the estimates $\hat{\Xi}_{t|t}$ computed by the Kalman filter \cite{Anderson:1979} 
to propagate on-line also sparse estimates $\hat{\Xi}_t$ of 
the equations which govern \eqref{StateMod2}.
Indeed, all the components in
$\hat{\Xi}_{t}$ whose absolute value is smaller than $\lambda$ turn out to be zero.
As discussed also later on, 
sparsity information can be interpreted as additional, noise-free measurements enforcing zero constraints on specific components of the state vector selected by $\lambda$. 
One can then prove that the estimates $\hat{\Xi}_t$ are Bayesian optimal if $\lambda$ correctly identifies the true sparsity pattern. 
More precisely, when the state-space is restricted to the non-zero parameter components and the model governing their evolution is correctly specified, no other unbiased estimator can achieve a lower estimation error variance. One can then state that\\

{\emph{if the sparsity information induced by $\lambda$ is correct, 
SKF returns the optimal (minimum variance) linear estimate.
Under Gaussian assumptions on the noises, the estimates are optimal among the set of all filters.}}\\

\noindent Furthermore, SKF generalizes Sindy to on-line settings:  as already mentioned, if system parameters do not vary in time the two procedures 
return the same sparse estimates under mild assumptions. This result, discussed in the next part of this Appendix, also shows that the Kalman filter propagation of posterior mean and covariance of system parameters maintains the information necessary to represent the entire family of possible sparse models that SINDy could produce.\\ 

Unknown parameters can be present in the filter,
e.g. they could enter the covariances $Q_t$ and represent switching instants where
$\Xi_t$ has an abrupt change.
Such parameters can be estimated 
by optimizing one of the many quality model indices 
computable by the Kalman filter. 
Examples are Generalized Cross Validation \cite{WahbaG:79,BottegalP17}, Marginal Likelihood \cite{Anderson:1979} or 
the predictivity score associated to the one-step ahead prediction error:
\begin{equation}\label{PredInd1}
\sum_{t} \ \big(y_t -  \Theta_t  \hat{\Xi}_{t|t-1}\big)^2,
\end{equation}
possibly using the innovation variances as weights which make such objective coincide with Marginal Likelihood. 
A simple strategy to exploit such criterion is to run a bank of filters in parallel and select the one which minimizes the objective \eqref{PredInd1}. 
Minimization of \eqref{PredInd1} plays a central role in system identification. According to \cite{Ljung:99}, as the number of data points tends to infinity, the parameter estimates obtained through prediction error minimization converge in probability to the parameter values that minimize the expected value of the squared prediction error. If the model structure is correctly specified (i.e., the true system lies within the model class), under some technical assumptions (such as model identifiability and Gaussian noise) this limiting parameter coincides with the true parameter. Furthermore,  the estimates are also asymptotically efficient and normally distributed, achieving the Cram\'er-Rao lower bound (smallest error variance inside the class of all unbiased estimators).\\
If the sparsity parameter $\lambda$ has to be estimated, 
the strategy devised in this paper is to optimize 
an objective similar to 
\eqref{PredInd1} where the one-step ahead prediction errors 
are calculated using the sparse estimates. 
Specifically, in our numerical examples, the parameter $\lambda$ is selected by minimizing the following prediction error objective:
\begin{equation}\label{PredInd2}
\sum_{t} \ \big(y_t - \Theta_t A \hat{\Xi}_{t-1}\big)^2.
\end{equation}
Grounded in the theoretical framework of prediction error minimization, this approach produces a parsimonious model that exhibits strong predictive accuracy and achieves asymptotic optimality as the sample size increases. If the true data-generating process is contained within the hypothesized model class, a parameter $\lambda$ inducing the correct sparsity pattern will be asymptotically selected. In fact, different choices of $\lambda$ lead to increased estimator variance, thereby inflating the prediction error and rendering those selections suboptimal in the limit.\\  
Additionally, an alternative online strategy could involve the use of validation data. A portion of the measurements $y_t$ may be withheld from the real-time data stream and used exclusively for validation purposes. Thanks to the flexible structure of the SKF, the algorithm can be queried at any desired time instant $t_k$ to produce sparse model estimates as a function of $\lambda$. The model corresponding to the highest predictive accuracy on the validation set can then be selected.

\subsection*{SKF properties}

Without sparsity on the state, assuming only that first- and second-order moments of $\Xi_t$ are correct,
Kalman filter theory ensures that $\hat{\Xi}_{t|t}$ is the minimum variance linear estimator.
In fact, the equations recursively project $\Xi_t$ onto the subspace generated
by all the outputs seen up to instant $t$. 
Using $\hat{\mathbb{E}}$ to denote projections, one thus has
$$
\hat{\Xi}_{t|t} = \hat{\mathbb{E}}[\Xi_t |  y_1,\ldots,y_t ].
$$
while $P_{t|t}$ is the covariance of the estimation error. 
If Gaussian assumptions on noises and initial condition also hold, 
the projection becomes the expectation $\mathbb{E}$ 
and $\hat{\Xi}_{t|t}$ is the minimum variance estimator.
This means that it minimizes the mean-squared-error (MSE)
over all the possible filters. In this Gaussian setting,
$P_{t|t}$ becomes the posterior covariance, i.e.
$$
P_{t|t} = Var[\Xi_t | y_1,\ldots,y_t ].
$$
Lines 10-13 are particularly important since they implement the sparsity module which resembles Sindy.
The sparsity parameter $\lambda$ defines the least important state components whose  indices are contained in 
$\text{smallinds}$. The estimate $\hat{\Xi}_{t|t}$ is then updated
through 
\begin{equation}\label{Line12}
 \hat{\Xi}_{t}=\hat{\Xi}_{t|t} - P_{t|t}(:,\text{smallinds}) \big(P_{t|t}(\text{smallinds},\text{smallinds})\big)^{-1} \hat{\Xi}_{t|t}(\text{smallinds}).
\end{equation}
This formula has a simple but important Bayesian interpretation and we start discussing it
still assuming Gaussianity. Let 
$z_t$ indicate  the Gaussian random variable $\Xi_t | y_1,\ldots,y_t$ discussed above, so that
$$
z_t \sim \mathcal{N}(\hat{\Xi}_{t|t},P_{t|t}).
$$
The set $\text{smallinds}$ then suggests the components of
$z_t$ to be set to zero. In Bayesian terms, this corresponds to considering
$z_t$ conditional on the event \emph{the components of $z_t$ contained in} $\text{smallinds}$ \emph{are null}.
Using standard formulas regarding estimation of jointly Gaussian vectors \cite{Anderson:1979}, 
the posterior mean $\hat{\Xi}_{t|t}$ 
is updated as follows

{{\small
\begin{eqnarray}\nonumber
\mathbb{E}[z_t | z_t(\text{smallinds} )=0] &=& \mathbb{E}[z_t] \\ \nonumber
&-& Cov\big(z_t,z_t(\text{smallinds} )\big)
\big(Var(z_t(\text{smallinds} )\big)^{-1} \mathbb{E}[z_t(\text{smallinds} )] \\ \nonumber
&=& \hat{\Xi}_{t|t} \\ \nonumber
&-& P_{t|t}(:,\text{smallinds}) \big(P_{t|t}(\text{smallinds},\text{smallinds})\big)^{-1} \hat{\Xi}_{t|t}(\text{smallinds})\\
\label{ProjLine12}
\end{eqnarray}
}}

and this, combined with \eqref{Line12} (Line 12 of SKF), implies that
$$
\hat{\Xi}_{t}=\mathbb{E}[z_t | z(\text{smallinds} )=0].
$$

It follows that, under Gaussian assumptions, SKF provides the minimum-variance sparse estimate, provided that $\lambda$ correctly identifies the sparsity pattern. More precisely, when the state-space is restricted to the non-zero parameter components and the model governing their evolution is correctly specified, no other unbiased estimator can achieve a lower error variance.\\ 
If Gaussian assumptions do not hold, \eqref{Line12} 
now implements a projection and we can replace  
$\mathbb{E}$ on the l.h.s. of \eqref{ProjLine12} with the symbol $\hat{\mathbb{E}}$.
So, in the absence
of Gaussianity, SKF remains optimal within the class of linear filters, assuming that $\lambda$ correctly specifies the sparsity
pattern.

\subsection*{Model predictivity indices}\label{Sec2}

Unknown parameters, also beyond $\lambda$, can be present in the Kalman filter. 
They can be estimated without resorting to any validation data 
by considering one of the many quality model indices 
made available by the Kalman filter. 
Examples include Generalized Cross Validation \cite{WahbaG:79,BottegalP17}, 
Marginal Likelihood \cite{Anderson:1979} or 
the predictivity score associated to the one-step ahead prediction error:
\begin{equation}\label{PredInd1}
\sum_{t} \ \big(y_t -  \Theta_t  \hat{\Xi}_{t|t-1}\big)^2.
\end{equation}
This important measure of model-predictivity returned in real-time by the Kalman filter 
has a rich and fruitful history in the identification field. 
A simple strategy is to run a bank of filters  in parallel 
containing different parameters and select the one which minimizes \eqref{PredInd1}.\\
If the sparsity parameter $\lambda$ has to be estimated, 
one can optimize 
a similar objective with the one-step ahead prediction errors 
calculated using the sparse estimates. 
Specifically, minimization w.r.t. $\lambda$ of
\begin{equation}\label{PredInd2}
\sum_{t} \ \big(y_t -  \Theta_t  A \hat{\Xi}_{t-1}\big)^2
\end{equation}
may suggest the sparse model with the largest predictive capability. 

\subsection*{Use of a non informative prior}\label{Sec3}

We have often mentioned the possible use of noninformative priors
to inizialize the filter. This corresponds to modeling the initial state
$\Xi_1$ as a zero-mean random vector with covariance $P=\gamma I$, then
letting $\gamma$ grow to infinity. Divergence of the prior variance  
indicates that no knowledge on the system state is available before observing the measurements.\\
In practice, one can exploit SKF equations with $\gamma$ set to a conveniently large 
value. A more elegant solution uses the matrix inversion lemma to 
reformulate Lines 5-7 as follows: 
\begin{eqnarray}\label{MinvL}
\hat{\Xi}_{t|t}&=& \hat{\Xi}_{t|t-1} + \big(\Theta_t^\top \Theta_t + \eta^2 P^{-1}_{t|t-1} \big)^{-1} \Theta_t^\top (y_t -  \Theta_t  \hat{\Xi}_{t|t-1})\\ \nonumber
P_{t|t} &=&  \eta^2 \big(\Theta_t^\top \Theta_t + \eta^2 P^{-1}_{t|t-1} \big)^{-1}.
\end{eqnarray}
When $t=1$, one has $\hat{\Xi}_{1|0}=\mu$ and $P^{-1}_{1|0}=P^{-1}$ .
To use a noninformative prior we set $\mu=0,P^{-1}=0$ and one obtains
\begin{eqnarray*}
\hat{\Xi}_{1|1}&=& \big(\Theta_1^\top \Theta_1 \big)^{\dagger} \Theta_1^\top (y_1 -  \Theta_1  \mu)\\
P_{1|1} &=&  \eta^2 \big(\Theta_1^\top \Theta_1 \big)^{\dagger}.
\end{eqnarray*}
where $\dagger$ denotes the pseudoinverse. 
The estimates can then be updated using \eqref{MinvL} with inverses replaced by pseudoinverses.
When $P(t|t)$ becomes full rank one can come back to exploit  Lines 5-7 of the SKF pseudocode.\\
Finally, consider the time-invariant case where $A=I,Q=0$ which implies $P_{t+1|t}=P_{t|t}$. 
If we use a noninformative prior with $\mu=0,P^{-1}=0$, from \eqref{MinvL} one can see that the estimate is independent of the noise variance so that one can set $\eta=1$.

\subsection*{Equivalence with Sindy in the time-invariant setting}\label{Sec4}

To discuss the relationship with Sindy, it is convenient to assume Gaussianity.
However, the same result stated below would be obtained replacing 
expectation $\mathbb{E}$ with projection $\hat{\mathbb{E}}$.
Gaussianity just simplifies the exposition.\\
Consider the time-invariant case where $A=I$ and $w_t$ is null for any $t$.
Thus, the state $\Xi_t$ does not vary in time,
and it is useful to let
$$
\Xi := \Xi_t.
$$ 
Without loss of generality, assume 
the the first $d_1$ components
of the Kalman estimate  $\hat{\Xi}_{t|t}$ of $\Xi$ are all and only those
with absolute value smaller than $\lambda$.
In addition, let us fix a generic instant $t$. So, also the dependence of the estimate on time can be omitted and
we define $$\hat{\Xi} := \hat{\Xi}_{t|t}$$ and partition such vector as follows
$$
\hat{\Xi}  = \left[\begin{array}{c} \hat{\Xi}_1 \\  \hat{\Xi}_2 \end{array}\right]
$$
with $\hat{\Xi}_1 \in \mathbb{R}^{d_1}$ and $\hat{\Xi}_2 \in \mathbb{R}^{d_2}$.
The unknown state vector is partitioned similarly, letting 
$$
\Xi = \left[\begin{array}{c} \Xi_1 \\  \Xi_2 \end{array}\right]
$$
with $\Xi_1 \in \mathbb{R}^{d_1}$ and $\Xi_2 \in \mathbb{R}^{d_2}$.
These two subvectors are given a noninformative prior: $\Xi_1$ and $\Xi_2$ are independent 
zero-mean Gaussian vectors of covariance $\gamma I_{d_1}$ and $\gamma I_{d_2}$
with $\gamma$ which will then grow to infinity.\\
The regression matrix at instant $t$ is denoted by $\Theta$ and $y$ is the vector with all the measurements up to instant $t$. So, 
the measurements model and the prior $\Xi$  is
\begin{equation}\label{MeasModStatic}
y = \Theta \Xi + e, \ \Xi \sim \mathcal{N}(0,P), \ P=\gamma I
\end{equation}
with the noise $e$ assumed of unit variance just to simplify notation.
It is also convenient to partition the regression matrix following what done before: 
$$
\Theta= \left[\begin{array}{cc} \Theta_a & \Theta_b \end{array}\right]
$$
where $\Theta_a$ contains the first $d_1$ columns of $\Theta$. Both $\Theta_a$ and $\Theta_b$ are assumed of full column rank.\\
 As already recalled, under Gaussian assumptions $P_{t|t}$ is the covariance matrix of
$\Xi$ conditional on the data $y$. Using \eqref{MeasModStatic} one obtains
\begin{eqnarray}\label{DefE0}
P_{t|t}&=& Var[\Xi|y]   \\ \nonumber
&=& P- P\Theta^\top \big(\Theta P \Theta^\top + I\big)^{-1} \Theta P
\\ \nonumber
&=& \big(\Theta^\top \Theta + P^{-1}\big)^{-1}.
\end{eqnarray}
When $\gamma$ grows to infinity one has
\begin{eqnarray}\label{DefE}
\lim_{\gamma \rightarrow +\infty} P_{t|t} &=& \big(\Theta^\top \Theta \big)^{-1} \\ \nonumber
&=:& \left[\begin{array}{cc} E_1 & E_{12} \\ E_{21} & E_2\end{array}\right]
\end{eqnarray}
where, again in accordance with the partitions introduced above, 
one has $E_1\in  \mathbb{R}^{d_1 \times d_1}$ and $E_2 \in \mathbb{R}^{d_2 \times d_2}$.\\
To study the relationship with Sindy, it is sufficient to investigate the output
$\hat{\Xi}^s$ obtained after one iteration of the sparsity module 
having $\hat{\Xi}$ as input 
(in the pseudo code, $\hat{\Xi}$ and $\hat{\Xi}^s$ are, respectively, $\hat{\Xi}_{t|t}$ and $\hat{\Xi}_{t}$).
The equivalence between Sindy and SKF in the time-invariant case corresponds to proving that, if $P$ is not informative, 
one has
$$
\hat{\Xi}^s = \left[\begin{array}{c} \hat{\Xi}^s_1 \\  \hat{\Xi}^s_2 \end{array}\right]
$$ 
where 
\begin{equation}\label{Xs1}
\hat{\Xi}^s_1 = 0_{d_1}
\end{equation}
(the null vector of dimension $d_1$),
while
\begin{equation}\label{Xs2}
\hat{\Xi}^s_2 = \big((\Theta_b)^\top \Theta_b\big)^{-1}(\Theta_b)^\top y,
\end{equation}
which corresponds to regression only onto
the remaining $d_2$ terms.
\eqref{Xs1} is immediately obtained just
evaluating \eqref{Line12} (Line 12 of the pseudocode) 
at $\text{smallinds}$.
On the other hand, using Bayesian arguments, we have already seen that
in any setting SKF sets to zero the components of the Kalman estimate with absolute
value smaller than $\lambda$.\\
The proof of \eqref{Xs2} will be obtained investigating the structure of
$\hat{\Xi}_2$ and then linking it to $\hat{\Xi}^s_2$.
It holds that
\begin{equation}\label{OP1}
\hat{\Xi}_2 = \mathbb{E}[\Xi_2 |y] = \mathbb{E}\big[ \mathbb{E}[\Xi_2 |\Xi_1, y] |y \big].
\end{equation}
Now, we decompose the inner projection into two oblique projections.
One has
\begin{equation}\label{OP2}
\mathbb{E}[\Xi_2 |\Xi_1, y] = \mathbb{E}_{|| \Xi_1}[\Xi_2 | y] + \mathbb{E}_{|| y}[\Xi_2 | \Xi_1].
\end{equation}
For what concerns the first term on the r.h.s. of \eqref{OP2}, one has
\begin{eqnarray}\label{FirstOP}
\mathbb{E}_{|| \Xi_1}[\Xi_2 | y] &=& Cov[\Xi_2,y | \Xi_1] \big(Var[y | \Xi_1]\big)^{-1}y \\ \nonumber
&=& Cov[\Xi_2, \Theta_b \Xi_2] \big(Var[\Theta_b\Xi_2 + e ]\big)^{-1}y \\ \nonumber
&=& \gamma (\Theta_b)^\top \big(\gamma \Theta_b (\Theta_b)^\top+ I \big)^{-1}y \\ \nonumber
&\stackrel{\text{$\gamma \rightarrow +\infty$}}{=}& \big((\Theta_b)^\top \Theta_b \big)^{-1} (\Theta_b)^\top y 
\end{eqnarray}
where the last equality is obtained letting $\gamma$ go to infinity.\\
For what concerns the second term on the r.h.s. of \eqref{OP2}, one has
\begin{eqnarray}\label{SecondOP}
\mathbb{E}_{|| y}[\Xi_2 | \Xi_1] &=& Cov[\Xi_2,\Xi_1 | y] \big(Var[\Xi_1 | y]\big)^{-1} \Xi_1 \\ \nonumber
&\stackrel{\text{$\gamma \rightarrow +\infty$}}{=}& E_{21} E_1^{-1} \Xi_1 
\end{eqnarray}
where the last equality holds using a non informative prior and the corresponding notation in \eqref{DefE}.
Finally, after projecting \eqref{FirstOP} and \eqref{SecondOP} onto $y$, we obtain the following expression for \eqref{OP1}:
$$
\hat{\Xi}_2 = \big((\Theta_b)^\top \Theta_b \big)^{-1} (\Theta_b)^\top y  + E_{21} E_1^{-1}\hat{\Xi}_1, 
$$
which implies
$$
\big((\Theta_b)^\top \Theta_b \big)^{-1} (\Theta_b)^\top y = \hat{\Xi}_2 - E_{21} E_1^{-1} \hat{\Xi}_1.
$$
The r.h.s. corresponds exactly to the r.h.s. of \eqref{Line12} 
evaluated  at $\verb"biginds"$, the set 
containing the indices $d_1+1,\ldots,d_1+d_2$.
We conclude that
$$
\Xi_2^s = \big(\Theta_b (\Theta_b)^\top \big)^{-1} (\Theta_b)^\top y 
$$
and this proves that {\bf in the time-invariant setting SKF and Sindy are equivalent if a non informative prior is assigned to $\Xi$}.\\
Finally, assume that the prior is informative. If $\Theta_k$ is the $k$th row of 
$\Theta$, at instant $t$ one has
$$
\Theta^\top \Theta= \sum_{k=1}^t \Theta^{\top}_k \Theta_k.
$$
Assume that for a certain index $\bar{t}$
\begin{equation}\label{Perst}
\frac{1}{\bar{t}} \sum_{k=1}^{\bar{t}} \Theta^{\top}_k \Theta_k \succ 0. 
\end{equation}
Then, for any (invertible) covariance prior $P$, as $t$ grows to infinity one has 
$$\frac{1}{t} \sum_{k=1}^t \Theta^{\top}_k \Theta_k+\frac{P^{-1}}{t} \approx \frac{1}{t} \sum_{k=1}^t \Theta^{\top}_k \Theta_k.$$
So, the effect of the prior $P$ on the Kalman estimates vanishes as time progresses and we come back to the situation
described at the end of Section
\ref{Sec3},
as if $\Xi$ were given a non informative prior. 
Overall, the arguments reported in this section lead to the following result.

\begin{proposition}
Consider the time-invariant setting where system parameters do not vary in time.
If the prior on the parameters is not informative (no knowledge on $\Xi$ is available before seeing the data) 
the estimates returned by SKF and Sindy perfectly match. If an informative prior is instead assigned to  
$\Xi$, SKF and Sindy are asymptotically equivalent if the condition in \eqref{Perst} holds. 
\end{proposition}

An important interpretation of this result is also that the sufficient statistics (posterior mean and covariances of system states) propagated by the Kalman filter encompass all possible sparse estimates that can be produced by SINDy.


\begin{thebibliography}{10}
\providecommand{\url}[1]{\texttt{#1}}
\expandafter\ifx\csname urlstyle\endcsname\relax
  \providecommand{\doi}[1]{doi:\discretionary{}{}{}#1}\else
  \providecommand{\doi}{doi:\discretionary{}{}{}\begingroup
  \urlstyle{rm}\Url}\fi

\bibitem{Astrom71}
K.~{\r{A}}str\"{o}m, P.~Eykhoff, System identification-a survey.
  \emph{Automatica} \textbf{7}, 123--162 (1971).

\bibitem{Soderstrom}
T.~S{\"o}derstr{\"o}m, P.~Stoica, \emph{System Identification} (Prentice-Hall)
  (1989).

\bibitem{Ljung:99}
L.~Ljung, \emph{System Identification - Theory for the User} (Prentice-Hall,
  Upper Saddle River, N.J.), 2nd ed. (1999).

\bibitem{Scholkopf01b}
B.~Sch\"{o}lkopf, A.~Smola, \emph{Learning with Kernels: Support Vector
  Machines, Regularization, Optimization, and Beyond}, Adaptive Computation and
  Machine Learning (MIT Press) (2001).

\bibitem{SpringerRegBook2022}
G.~Pillonetto, T.~Chen, A.~Chiuso, G.~{De Nicolao}, L.~Ljung, \emph{Regularized
  System Identification} (Springer) (2022).

\bibitem{PillonettoPNAS}
G.~Pillonetto, L.~Ljung, Full {B}ayesian identification of linear dynamic
  systems using stable kernels. \emph{Proceedings of the National Academy of
  Sciences USA} \textbf{120} (2023).

\bibitem{Tibshirani:96}
R.~Tibshirani, Regression shrinkage and selection via the {L}asso.
  \emph{Journal of the Royal Statistical Society: Series B (Statistical
  Methodology)} \textbf{58}, 267--288 (1996).

\bibitem{Bai2019}
E.~Bai, C.~Cheng, W.~Zhao, Variable selection of high-dimensional
  non-parametric nonlinear systems by derivative averaging to avoid the curse
  of dimensionality. \emph{Automatica} \textbf{101}, 138--149 (2019).

\bibitem{Bai2018}
B.~{Mu}, W.~{Zheng}, E.~{Bai}, Variable selection and identification of
  high-dimensional nonparametric additive nonlinear systems. \emph{IEEE
  Transactions on Automatic Control} \textbf{62}, 2254--2269 (2017).

\bibitem{Rosasco2013}
L.~Rosasco, S.~Villa, S.~Mosci, M.~Santoro, A.~Verri, Nonparametric sparsity
  and regularization. \emph{Journal of Machine Learning Research} \textbf{14},
  1665--1714 (2013).

\bibitem{Smith2014}
R.~Smith, Frequency domain subspace identification using nuclear norm
  minimization and {H}ankel matrix realizations. \emph{IEEE Transactions on
  Automatic Control} \textbf{59}, 2886--2896 (2014).

\bibitem{Stoddard2017}
J.~{Stoddard}, J.~{Welsh}, H.~{Hjalmarsson}, {EM}-based hyperparameter
  optimization for regularized volterra kernel estimation. \emph{IEEE Control
  Systems Letters} \textbf{1}, 388--393 (2017).

\bibitem{Champion2020}
K.~Champion, P.~Zheng, A.~Aravkin, S.~Brunton, J.~Kutz, A unified sparse
  optimization framework to learn parsimonious physics-informed models from
  data. \emph{IEEE Access} \textbf{8}, 169259--169271 (2020).

\bibitem{Karniadakis2021}
G.~Karniadakis, \emph{et~al.}, Physics-informed machine learning. \emph{Nature
  Reviews Physics} \textbf{3}, 422--440 (2021).

\bibitem{nghiem2023}
T.~Nghiem, \emph{et~al.}, Physics-informed machine learning for modeling and
  control of dynamical systems, in \emph{Proceedings of the American Control
  Conference} (2023), pp. 3735--3750.

\bibitem{Raissi2019}
M.~Raissi, P.~Perdikaris, G.~Karniadakis, Physics-informed neural networks: A
  deep learning framework for solving forward and inverse problems involving
  nonlinear partial differential equations. \emph{Journal of Computational
  Physics} \textbf{378}, 686--707 (2019).

\bibitem{tibshirani_regression_1996}
R.~Tibshirani, Regression shrinkage and selection via the {LASSO}.
  \emph{Journal of the Royal Statistical Society. Series B (Methodological)}
  pp. 267--288 (1996).

\bibitem{PC08}
T.~Park, G.~Casella, The {B}ayesian lasso. \emph{Journal of the American
  Statistical Association} \textbf{103}, 681--686 (2008).

\bibitem{Daub2004}
I.~Daubechies, M.~Defrise, C.~De~Mol, An iterative thresholding algorithm for
  linear inverse problems with a sparsity constraint. \emph{Communications on
  Pure and Applied Mathematics} \textbf{57}, 1413--1457 (2004).

\bibitem{Brunton2016}
S.~Brunton, J.~Proctor, J.~Kutz, Discovering governing equations from data by
  sparse identification of nonlinear dynamical systems. \emph{Proceedings of
  the National Academy of Sciences} \textbf{113}, 3932--3937 (2016).

\bibitem{Champion2019}
K.~Champion, B.~Lusch, J.~Kutz, S.~Brunton, Data-driven discovery of
  coordinates and governing equations. \emph{Proceedings of the National
  Academy of Sciences} \textbf{116}, 22445--22451 (2019).

\bibitem{kalman1960}
R.~Kalman, A new approach to linear filtering and prediction problems.
  \emph{Journal of basic Engineering} \textbf{82}, 35--45 (1960).

\bibitem{Chui2017}
C.~Gui, G.~Chen, \emph{Kalman Filtering with real-time applications} (Springer)
  (2017).


\bibitem{DeVivo2014}
A.~De~Vivo, J.~Leofanti, Vega in-flight modal identification with the
  operational modal analysis technique. \emph{Journal of Spacecraft and
  Rockets} \textbf{51} (2014).

\bibitem{Toth2010}
R.~Toth, \emph{Modeling and identification of linear parameter-varying systems}
  (Springer) (2010).

\bibitem{Sontag1996}
E.~Sontag, Interconnected automata and linear systems: A theoretical framework
  in discrete-time, in \emph{Hybrid Systems III} (Springer Berlin Heidelberg),
  vol. 1066 of \emph{Lecture Notes in Computer Science}, pp. 436--448 (1996).

\bibitem{JuloskiLN2005}
A.~Juloski, \emph{et~al.}, Comparison of four procedures for the identification
  of hybrid systems, in \emph{Hybrid Systems: Computation and Control}, Lecture
  Notes in Computer Science (Springer Berlin Heidelberg), pp. 354--369 (2005).

\bibitem{PillonettoHybrid}
G.~Pillonetto, A new kernel-based approach to hybrid system identification.
  \emph{Automatica} \textbf{70}, 21--31 (2016).

\bibitem{Yan2023}
H.~Yan, F.~Zhang, J.~Wang, Thermodynamic and dynamical predictions for
  bifurcations and non-equilibrium phase transitions. \emph{Communications
  Physics} \textbf{6} (2023).

\bibitem{Rhind2012}
N.~Rhind, P.~Russell, \emph{Signaling Pathways that Regulate Cell Division}
  (Cold Spring Harbor Perspectives in Biology) (2012).

\bibitem{Cadart2018}
C.~Cadart, \emph{et~al.}, Size control in mammalian cells involves modulation
  of both growth rate and cell cycle duration. \emph{Nature Communications}
  \textbf{9} (2018).

%

\bibitem{Holmes1983}
P.~Holmes, J.~Guckenheimer, Nonlinear oscillations, dynamical systems, and
  bifurcations of vector fields. \emph{Applied Mathematical Sciences}
  \textbf{42} (1983).

\bibitem{Hassard1981}
B.~Hassard, N.~Kazarinoff, Y.~Wan, \emph{Theory and Applications of Hopf
  Bifurcation} (Cambridge University Press) (1981).

\bibitem{Boyd1985}
S.~{Boyd}, L.~{Chua}, Fading memory and the problem of approximating nonlinear
  operators with {V}olterra series. \emph{IEEE Transactions on Circuits and
  Systems} \textbf{32}, 1150--1161 (1985).

\bibitem{And1985}
P.~Andersson, Adaptive forgetting in recursive identification through multiple
  models. \emph{Int. J. Control} \textbf{42}, 1175--1193 (1985).

\bibitem{Gust2001}
F.~Gustafsson, \emph{Adaptive Filtering and Change Detection} (Wiley, New York)
  (2001).

\bibitem{Ohl2010}
H.~Ohlsson, L.~Ljung, S.~Boyd, Segmentation of arx-models using sum-of-norms
  regularization. \emph{Automatica} \textbf{46}, 1107--1111 (2010).

\bibitem{Sohn2003}
H.~Sohn, \emph{et~al.}, \emph{A review of structural health monitoring
  literature: 1996--2001}, Tech. Rep. LA-13976-MS, Los Alamos National
  Laboratory (2003).

\bibitem{Zonta2004}
D.~Zonta, M.~Chaabane, M.~De~Stefano, C.~Christopoulos, Damage detection and
  localization in a steel frame using stochastic subspace identification.
  \emph{Structural Safety} \textbf{26}~(3), 291--306 (2004).

\bibitem{Fassois2007}
S.~D. Fassois, J.~S. Sakellariou, Time-series methods for fault detection and
  identification in vibrating structures. \emph{Phil. Trans. R. Soc. A}
  \textbf{365}~(1851), 411--448 (2007).

\bibitem{Wahba:90}
G.~Wahba, \emph{Spline models for observational data} (SIAM, Philadelphia)
  (1990).

\bibitem{deBoor2001}
C.~de~Boor, \emph{A practical guide to splines} (Springer, New York) (2001).

\bibitem{DurbinKoopman2012}
J.~Durbin, S.~J. Koopman, \emph{Time Series Analysis by State Space Methods}
  (Oxford University Press), 2nd ed. (2012).

\bibitem{Pillow2008}
J.~W. Pillow, \emph{et~al.}, Spatio-temporal correlations and visual signalling
  in a complete neuronal population. \emph{Nature} \textbf{454}~(7207),
  995--999 (2008).

\bibitem{Maine1985}
R.~Maine, K.~Iliff, \emph{Identification of dynamic systems, theory and
  formulation}, Tech. Rep. RP-1138, NASA (1985).

\bibitem{Jat2015}
R.~Jategaonkar, \emph{Flight Vehicle System Identification: A Time-Domain
  Methodology} (AIAA) (2015).

\bibitem{Majeed2018}
M.~Majeed, D.~Vikalp, Aircraft neural modeling and parameter estimation using
  neural partial differentiation. \emph{Aircraft engineering and aerospace
  technology} \textbf{90}, 764--778 (2018).

\bibitem{Amin2019}
B.~Amin, Flight dynamics modeling of elastic aircraft using signal
  decomposition methods. \emph{Proceedings of the Institution of Mechanical
  Engineers, Part G: Journal of Aerospace Engineering} \textbf{233}, 4380--4395
  (2019).

\bibitem{Wu2021}
W.~Wu, R.~Chen, An improved on-line system identification method for tiltrotor
  aircraft. \emph{Aerospace Science and Technology} \textbf{110}, 106491
  (2021).

\bibitem{Verma_Peyada_2021}
H.~Verma, N.~Peyada, Estimation of aerodynamic parameters near stall using
  maximum likelihood and extreme learning machine-based methods. \emph{The
  Aeronautical Journal} \textbf{125}, 489--509 (2021).

\bibitem{Greenburg1951}
H.~Greenburg, \emph{A survey of methods for determining longitudinal-stability
  derivatives of an airplane from transient flight data}, Tech. Rep. TN-2340,
  NACA (1951).

\bibitem{Shinbrot1951}
M.~Shinbrot, \emph{A least squares curve fitting method with applications to
  the calculate of stability coefficients from transient-response data}, Tech.
  Rep. TN-23, NACA (1951).

\bibitem{Haiquan2022}
L.~Haiquan, C.~Xiaoqian, The research on the stochastic dynamics of the
  aircraft flight attitude. \emph{Acta Aeronautica et Astronautica Sinica}
  \textbf{43} (2022).

\bibitem{Li2023}
H.~Li, \emph{et~al.}, Identification of a stochastic dynamic model for aircraft
  flight attitude based on measured data. \emph{International Journal of
  Aerospace Engineering} \textbf{2023} (2023).

\bibitem{Strogatz2015}
S.~Strogatz, \emph{Nonlinear Dynamics and Chaos: With Applications to Physics,
  Biology, Chemistry, and Engineering} (CRC Press) (2015).

\bibitem{Anderson:1979}
B.~Anderson, J.~Moore, \emph{Optimal Filtering} (Prentice-Hall, Englewood
  Cliffs, N.J., USA) (1979).

\bibitem{WahbaG:79}
G.~Golub, M.~Heath, G.~Wahba, Generalized cross-validation as a method for
  choosing a good ridge parameter. \emph{Technometrics} \textbf{21}, 215--223
  (1979).

\bibitem{BottegalP17}
G.~Bottegal, G.~Pillonetto, The generalized cross validation filter.
  \emph{Automatica} \textbf{90}, 130--137 (2017).

\end{thebibliography}
\end{document}